# AI-integrated models for assessing agricultural resilience

Joshua R. Waite[1], Dana Golden[2], Brett Indelicato[2], Kevin Camp[3], Mojdeh Saadati[5], Shannon Regan[4], Patrick Schnable[6], Baskar Ganapathysubramanian[1,7], Carlos Messina[4], Suzanne Thornsbury[3*], Soumik Sarkar[1,7*]

[1*]Translational AI Center, Iowa State University, Ames, IA, USA.

[2] Department of Economics, Stony Brook University, Stony Brook, NY, USA.

[3] Department of Food and Resource Economics, University of Florida, Gainesville, FL, USA.

[4]Department of Horticultural Sciences, University of Florida, Gainesville, FL, USA.

[5]Department of Computer Science, Iowa State University, Ames, IA, USA.

[6]Department of Agronomy, Iowa State University, Ames, IA, USA.

[7]Department of Mechanical Engineering, Iowa State University, Ames, IA, USA.

*Corresponding author(s). E-mail(s): thornsbs@ufl.edu; soumiks@iastate.edu
Contributing authors: jrwaite@iastate.edu; dana.golden@stonybrook.edu; soumiks@iastate.edu; brett.indelicato@stonybrook.edu; kevincamp@ufl.edu; msaadati@iastate.edu; shannonregan@ufl.edu; schnable@iastate.edu; baskarg@iastate.edu; cmessina@ufl.edu

**Keywords:** Agricultural Resilience, Agentic AI, Social-Ecological Systems, Biophysical Modeling, Economic Modeling

**Abstract:** Agricultural supply chains are vulnerable to disruptions through linked biophysical and economic systems. We develop an AI-powered tool that integrates economic models (GTAP) with biophysical models (APSIM) to analyze supply chain shocks, enabling policymakers and market participants to assess cross-disciplinary impacts through queries and responses written in natural language.

## Introduction

Collective ongoing pressures, sometimes labeled a "polycrisis", (Rakowski et al., 2025) have exposed both the robustness and the fragility of agricultural systems. In most cases, some production has continued with adjusted trade flows over time, but recent disruptions revealed previously invisible systemic risks including dependence on narrow logistical corridors, concentrated processing capacity, and limited slack in labor and input markets (Lazebnik, 2025; Manfredo et al., 2025). Sustainability of this complex social-ecological system (SES) requires combining biophysical, economic, and systemic-risk perspectives in ways that exceed any single discipline (Batie, 2008). We define agricultural resilience as the ability of agricultural systems to withstand and adjust to exogenous shocks without persistent losses in output, income, or ecological function (Gardner et al., 2019; Urruty et al., 2016). Because dynamic feedback loops link physical production to economic incentives and trade, integrated models are required to adequately assess supply chain vulnerability (Stone and Rahimifard, 2018).

Currently, no technical framework integrates biophysical and economic agricultural models automatically. Prior work has advanced biophysical modeling (Lwin et al., 2024) and economic analysis (Korder et al., 2024; Ahumadia and Villalobos, 2009) largely in disciplinary isolation, with some recent efforts beginning to bridge these perspectives (Lazebnik 2025; Liu et al., 2025). Agentic AI systems, which can interpret queries in natural-language, route the questions or instructions to specialized tools, and coordinate the information flow between models (Yao et al., 2022; Wu et al., 2024), offer a path toward model integration. These techniques have not yet been adapted to social–ecological systems where feedback among human decision-making, physical constraints, and institutional structure is central.

We present a conceptual vulnerability framework to categorize the agricultural supply chain as an SES delineating biophysical and economic exposures into discrete components we can model and a working prototype that couples domain models through a lightweight AI orchestration layer. Our conceptual framework considers feedback loops between physical and economic factors, separating economic risks into sectoral, supply chain network, and systemic macroeconomic components (Fig. 1). The framework does not replace human analysts but provides an automated first-level assessment of the fundamentals of the system and their interactions. Agriculture is a critical global SES, subject to multiple sources of biophysical and economic risk. The agriculture sector combines layers of complexity and is data-rich, with many pre-existing models that agentic AI can synthesize into a coherent system. Thus, we focus on agriculture as a prototype that can be generalized to other multifaceted SES.

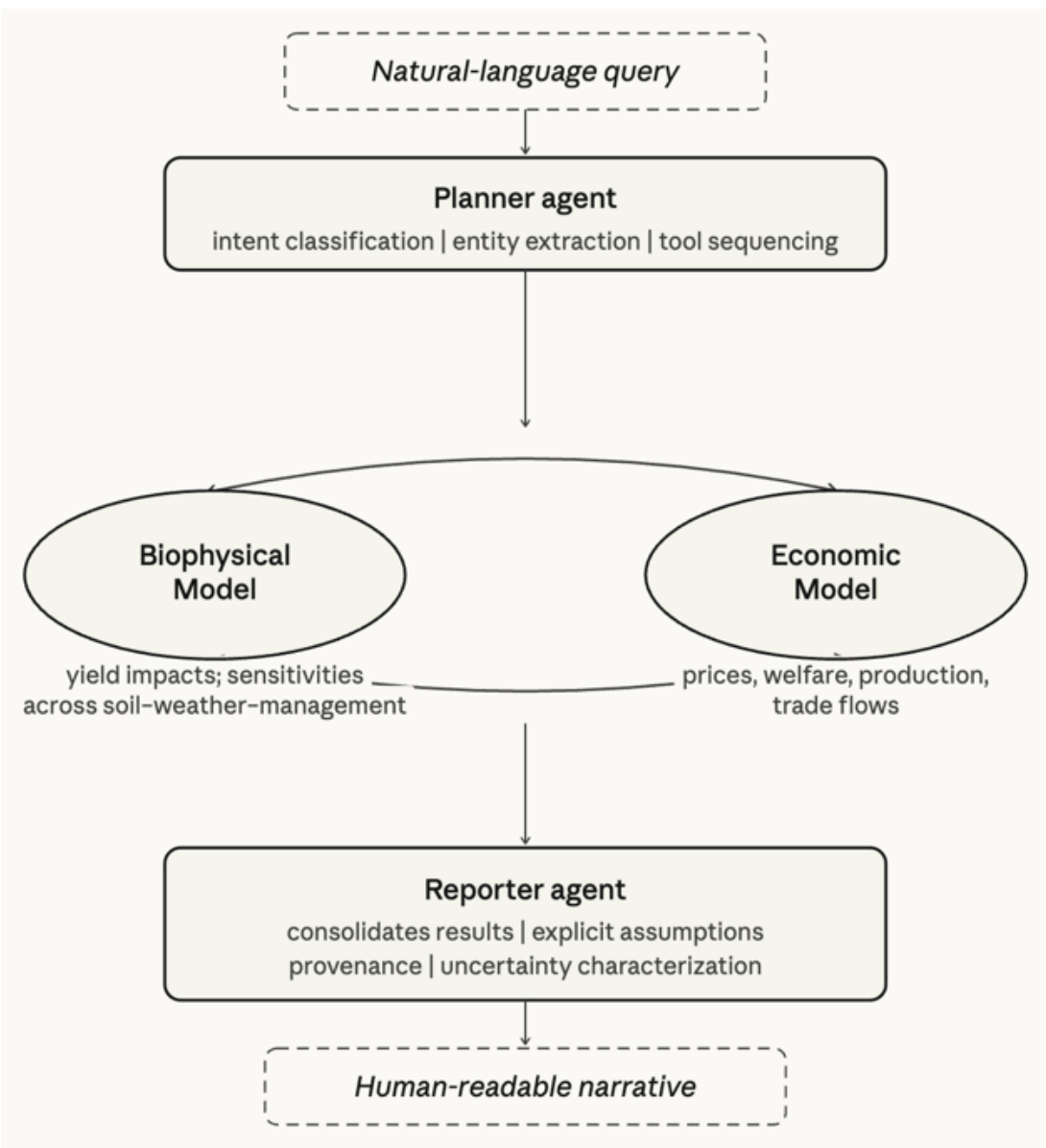

**Fig. 1** Overview of the agentic AI framework. The user poses a natural-language query; the planner agent parses it, selects and sequences the appropriate domain models, and the reporter agent synthesizes results into a unified response.

## Methods

We implement a two-agent architecture: a planner agent (orchestrator) and a reporter agent (synthesis). The planner receives a natural-language query and performs intent classification, entity extraction, model selection and sequencing. Tools are invoked through a common interface with typed inputs and outputs and validation checks. Requests that do not benefit from quantitative modeling are routed to a standard LLM constrained to document-based responses.

For biophysical tool calls, the planner assembles Agricultural Production Systems sIMulator (APSIM) emulator inputs (e.g. weather perturbation, soil, management, and genetic settings) and requests predicted outcomes with uncertainty estimates and sensitivity information. The APSIM emulator follows a three-stage pipeline. First, a large ensemble of maize simulations is generated using APSIM across diverse combinations of weather, soil, genetics, and management. Second, a neural network is trained to learn the input–output mapping of these simulations, approximating APSIM's behavior at dramatically reduced computational cost. Third, the trained emulator is deployed as a surrogate that accepts user-defined inputs and returns yield predictions with predictive uncertainty.

The biophysical foundation rests on process-based crop models. These explicitly represent the mechanisms governing crop development, including integrating genetics, management practices, and weather, to predict yields, biomass accumulation, and nutrient dynamics (Holzworth et al., 2014; Jones et al., 2003). When properly calibrated, such models enable counterfactual scenario analysis that extends beyond historical observations while retaining physiological interpretability.

However, existing high-fidelity models are computationally expensive and large-ensemble or repeated scenario analyses can be prohibitive. To address the computational limitations, we utilize a neural network emulator trained in extensive APSIM simulations that span various soil, climate, genetic, and management conditions (Saadati et al., 2026). The emulator provides orders-of-magnitude speedups while preserving predictive fidelity and, crucially, is differentiable. Differentiability enables gradient-based sensitivity analysis to identify leverage points within integrated frameworks (Franke et al., 2020; Shahhosseini et al., 2021). Coordinated intercomparisons, such as AgMIP and GGCMI have advanced ensemble-based risk quantification globally (Rosenzweig et al., 2013; Rosenzweig et al., 2014), and the emulator builds on this tradition to enable high-throughput coupling with economic models.

For economic tool calls, the planner converts scenario statements into shock specifications (e.g. sector productivity changes, trade-cost adjustments, tariff shocks) and requests relevant outputs (e.g. prices, quantities, welfare, bilateral trade flows).

The economic foundation captures incentives, institutional structures, and interactions of agents across sectors and regions. Traditional economic modeling has focused on efficient systems with minimal slack and long supply chains that exploit economies of scale; while these designs have lowered food prices, they have also concentrated vulnerability. The COVID-19 pandemic and the Russia–Ukraine war exposed these fragilities in global agricultural supply chains. Supply chain weakness was particularly strong for livestock and Black Sea winter wheat (Lwin et al., 2024; Korder et al., 2024). This led to a contemporary focus on resilience and the tradeoff between robustness and efficiency (Lazebnik 2025; Manfredo et al., 2025; Xue et al., 2025).

We employ two complementary economic modeling approaches. Supply chain network analysis traces goods from upstream producers through distributors to consumers, quantifying dependence on specific nodes, and evaluating resilience to disruption. A computable general equilibrium (CGE) framework is necessary to trace how shocks propagate through the broader economy via cross-commodity substitution, international competition, and sectoral feedback loops. These mechanisms cannot be fully captured by partial-equilibrium methods. We use Global Trade Analysis Project (GTAP) for macroeconomic and trade modeling that represents worldwide trade flows through neoclassical micro-foundations and a multilateral trade database (Christopher and Peck, 2004).

Cross-domain coupling is achieved by passing summarized APSIM outputs (e.g., yield changes by region) as structured inputs to GTAP shock generation, or vice versa. Data inputs can be obtained via APIs designed to feed APSIM and GTAP. In the current implementation, these inputs are hard-coded, but the interfaces support substitution with live pipelines.

Our framework couples chosen domain models through a two-agent AI architecture. A planner agent receives a natural-language query, performs intent classification and entity extraction (crops, regions, time horizon, shock type), and sequences tool calls. For biophysical sub-questions, the planner routes inputs to the APSIM emulator to evaluate yield impacts and compute sensitivities across soil–weather–management configurations. For macroeconomic sub-questions, it configures GTAP shocks (productivity changes, trade costs, tariffs) and retrieves equilibrium outcomes including prices, welfare, production, and trade flows.

When a query spans both domains, the planner passes structured outputs from one model as inputs to the other. The reporter agent then consolidates all results into a human-readable narrative with explicit assumptions, provenance, and uncertainty characterization. Conceptual queries (definitions, interpretation, and qualitative background) are handled by a bounded language-model interface without invoking domain simulators.

## Results

To illustrate our approach, we analyze a hypothetical tariff on fertilizer imports from Canada (Fig. 2). An analyst poses the question: What would a 20% tariff on Canadian fertilizer exports to the United States imply for corn yields in Iowa? In 2024 Canada supplied approximately 50% of U.S. fertilizer imports by value, including 85% of potassium and 30% of nitrogen. Iowa is the largest U.S. corn production state supplying 17% of national corn volume that same year. In our framework, the planner agent interprets the chosen event as a macroeconomic shock with biophysical consequences and routes the query first to GTAP, where the tariff can be implemented as an increase in trade costs. GTAP is used to estimate equilibrium price adjustments, import volumes, and the degree of substitution towards alternative suppliers or domestic production, quantifying the rise in effective input costs faced by U.S. producers. For the trade shock, the system applied a 1.2 tariff factor to chemical products (the GTAP sector that includes fertilizers) imported from Canada. The resulting computable general equilibrium (CGE) simulation projected declines in U.S. GDP, a reduction in aggregate U.S. imports, and decreases in domestic chemical and fertilizer consumption.

The planner agent then translates these economic outputs into agronomically meaningful inputs for APSIM. Specifically, the overall decrease in fertilizer consumption results in reductions in application rates by Iowa corn producers consistent with the equilibrium solution. These altered inputs are passed to the APSIM emulator The emulator simulates yield responses under reduced nitrogen availability across representative soils, climates, and management regimes in the target region (defined in the initial query) and returns yield impacts and sensitivities, identifying where production losses are most pronounced and the agronomic factors that moderate or amplify fertilizer constraints. For example, the baseline nitrogen-at-sowing was adjusted to reflect a reduction in fertilizer application per hectare. Under representative Iowa environmental conditions, the APSIM emulator projected a reduction in total corn grain weight ($g/m^2$), which is indicative of yield loss

The planner agent checks that tool execution is complete (verification step) and then calls the reporter agent with tool outputs, metadata (assumptions, parameter choices, run identifiers), and the original question. The reporter agent synthesizes these results into a unified narrative linking trade policy with fertilizer markets, crop productivity, and resilience. The reporter produces a response that answers the query, traces each tool's contribution, lists assumptions and scope, and highlights uncertainties. This scenario demonstrates how the framework translates a macro-level trade policy shift into a localized, field-level agronomic outcome without manual data harmonization between models. An example of the synthesized analytical output is shown in Fig. 2. The scenario assumes no immediate policy retaliation and no short-run rationing beyond price-mediated effects.

Our example demonstrates the value of coordinated cross-domain modeling: fertilizer tariffs affect yields not directly, but through price signals, substitution patterns, and producer responses. The mechanisms of impact span biophysical and economic domains. By coordinating existing specialized models within a single workflow, the framework makes these linkages and feedback loops explicit and enables rapid exploration of how trade policies translate into on-farm outcomes. We show that new models are not necessary to bridge the gap; but that synchronization of existing state-of-the-art models can provide the benefits of flexible, multi-domain modeling without new model development.

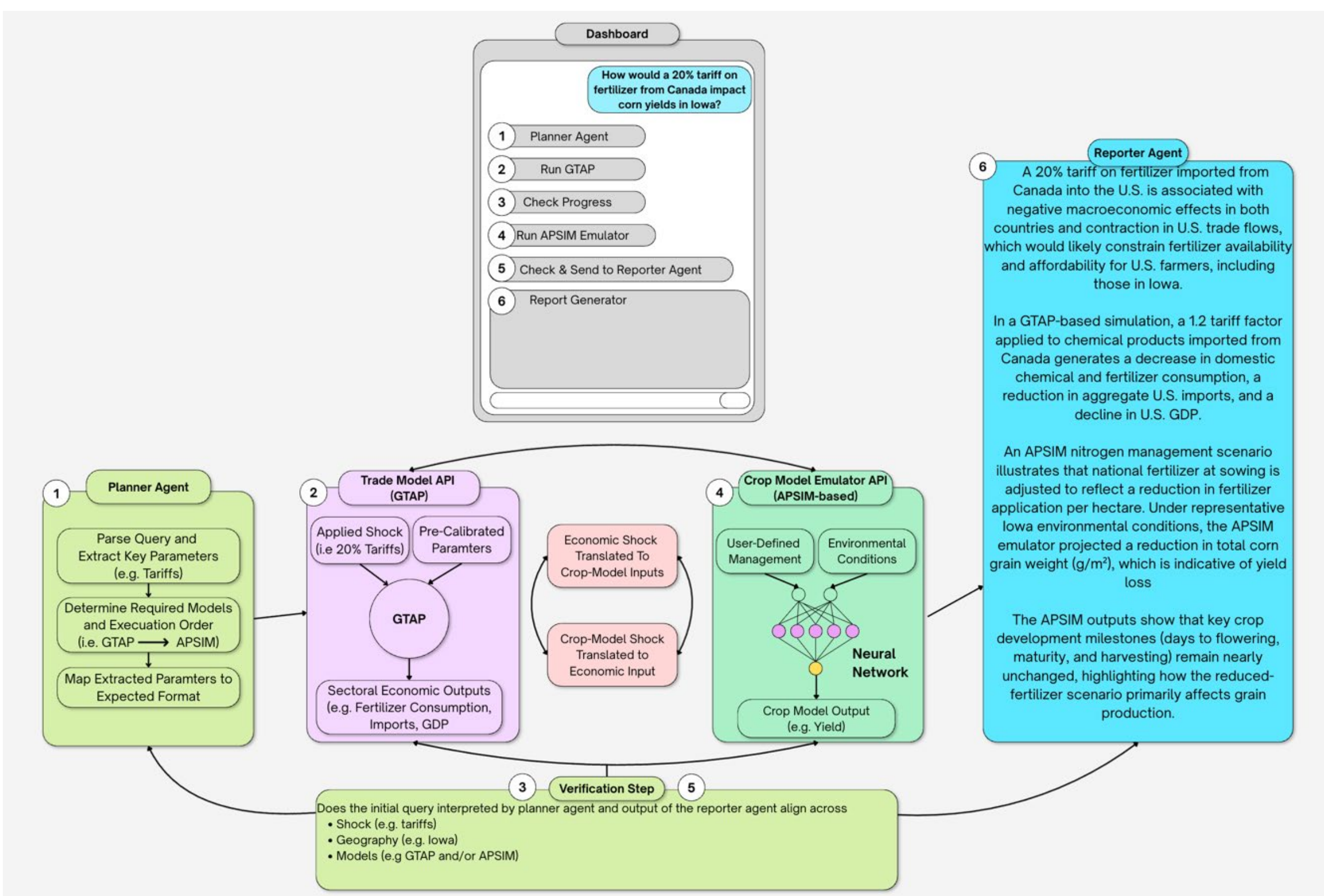

**Fig. 2** Technical architecture and illustrative scenario. A fertilizer-tariff shock is routed through GTAP and then the APSIM emulator, demonstrating cross-domain coupling.

## Discussion

The framework inherits limitations from its components. Mapping economic outcomes to agronomic decisions introduces additional structural assumptions, which we document explicitly but do not eliminate. APSIM emulation depends on the coverage and calibration of the simulation ensemble and does not capture behavioral adaptation beyond the changes in management parameters. GTAP captures equilibrium adjustments but abstracts from short-run frictions, credit constraints, and distributional heterogeneity among producers. The agentic architecture is designed to accommodate additional models and agents as these components are developed.

In advancing the integration of the biophysical (APSIM) and macroeconomic (GTAP) models, discrepancies in definitions, parameters, and scale currently prevent reportable quantified results. For example, in GTAP, fertilizers for corn production fall under the chemicals and fertilizers sector, meaning a tariff on all Canadian chemical and fertilizer imports would overestimate the economic responses to a fertilizer-specific tariff. In APSIM, only nitrogen at sowing is currently parameterized in the emulator, which underestimates the effects of a tariff on fertilizer blends requiring nitrogen, phosphorus, and potassium.

We do not yet model strategic interactions among decision-makers. Obviously, this presents an important limitation when analyzing trade wars, where best-response functions and retaliatory dynamics shape outcomes. Strategic-interaction modules are a natural extension of our architecture. Future work will incorporate empirically updated demand elasticities, dynamic adjustment paths, and explicit input-substitution modeling and could extend coverage to additional inputs such as water, labor, and energy.

## Conflict of Interest

*The authors declare that the research was conducted in the absence of any commercial or financial relationships that could be construed as a potential conflict of interest.*

## Author Contributions

JW: Formal Analysis, Methodology, Software, Visualization, Writing – original draft, Writing – review & editing; DG : Formal Analysis, Methodology, Software, Visualization, Writing – original draft, Writing – review & editing; BI: Methodology, Visualization, Writing – original draft, Writing – review & editing; KC: Methodology, Visualization, Writing – review & editing; MS: Formal Analysis, Methodology; SR: Writing – review & editing; PS: Conceptualization, Funding Acquisition; BG: Conceptualization, Funding Acquisition, Supervision; CM: Conceptualization, Funding Acquisition, Supervision; ST: Conceptualization, Funding Acquisition, Supervision, Validation, Writing – original draft, Writing – review & editing; SS: Conceptualization, Funding Acquisition, Supervision

## Funding

This effort was sponsored in whole or in part by the United State Government (USG). The U.S. Government is authorized to reproduce and distribute original submissions for publication for Governmental purposes notwithstanding any copyright notation thereon. In part, this material is based upon work supported by NSF under Grant NRT-HDR 2125295.

**Data Availability Statement**

The datasets used for this study are available from the corresponding author upon reasonable request.